\documentclass[12pt, draftclsnofoot, onecolumn]{IEEEtran}
\IEEEoverridecommandlockouts
\usepackage{amsfonts}
\usepackage{amsmath}
\usepackage{amssymb}
\usepackage{urwchancal}
\usepackage{algorithm}
\usepackage{algorithmicx}
\usepackage{algpseudocode}
\usepackage{dsfont}
\usepackage{authblk}
\usepackage{bbm}
\usepackage{graphicx}
\usepackage{epstopdf}
\usepackage{subfigure}
\usepackage{stfloats}
\usepackage{eqnarray}
\usepackage{makecell}
\usepackage{multirow}
\usepackage{enumerate}
\usepackage{booktabs}
\usepackage{cite}
\usepackage{balance}
\usepackage{color}
\usepackage{bm}
\usepackage{caption}
\usepackage{mathtools}
\usepackage{url}

\DeclarePairedDelimiter\autobracket{(}{)}
\DeclarePairedDelimiter\autosquare{[}{]}
\newcommand{\br}[1]{\autobracket*{#1}}
\newcommand{\sq}[1]{\autosquare*{#1}}

\newcommand{\ab}[1]{\left|#1\right|}

\begin{document}
\title{\LARGE Secure Distributed On-Device Learning Networks With Byzantine Adversaries}

\author{
\mbox{\small Yanjie Dong,~\IEEEmembership{\small Student Member, IEEE},}
\mbox{\small Julian Cheng,~\IEEEmembership{\small Senior Member, IEEE},}
\mbox{\small Md. Jahangir Hossain,~\IEEEmembership{\small Senior Member, IEEE},}
\mbox{\small and Victor C. M. Leung,~\IEEEmembership{\small Fellow, IEEE}}
\thanks{
This work was supported in part by a UBC Four-Year Doctoral Fellowship, in part by the Natural Science and Engineering Research Council of Canada, and in part by the National Engineering Laboratory for Big Data System Computing Technology at Shenzhen University, China. (Corresponding author: Victor C. M. Leung.)}
\thanks{Y. Dong is with the Department of Electrical and Computer Engineering, The University of British Columbia, Vancouver, BC V6T 1Z4, Canada (email:ydong16@ece.ubc.ca).}
\thanks{J. Cheng and M. J. Hossain are with the School of Engineering, The University of British Columbia, Kelowna, BC V1V 1V7, Canada (email:\{julian.cheng, jahangir.hossain\}@ubc.ca).}
\thanks{V. C. M. Leung is with the College of Computer Science and Software Engineering, Shenzhen University, Shenzhen 518060, China, and the Department of Electrical and Computer Engineering, The University of British Columbia, Vancouver, BC \mbox{V6T 1Z4}, Canada (e-mail: vleung@ieee.org).}
}

\maketitle
\begin{abstract}
The privacy concern exists when the central server has the copies of datasets.
Hence, there is a paradigm shift for the learning networks to change from centralized in-cloud learning to distributed \mbox{on-device} learning.
Benefit from the parallel computing, the on-device learning networks have a lower bandwidth requirement than the in-cloud learning networks.
Moreover, the on-device learning networks also have several desirable characteristics such as privacy preserving and flexibility.
However, the \mbox{on-device} learning networks are vulnerable to the malfunctioning terminals across the networks.
The worst-case malfunctioning terminals are the Byzantine adversaries, that can perform arbitrary harmful operations to compromise the learned model based on the full knowledge of the networks.
Hence, the design of secure learning algorithms  becomes an emerging topic in the on-device learning networks with Byzantine adversaries.
In this article, we present a comprehensive overview of the prevalent secure learning algorithms for the two promising  on-device learning networks: Federated-Learning networks and decentralized-learning networks.
We also review several future research  directions in the \mbox{Federated-Learning} and decentralized-learning networks.
\end{abstract}

\section{Introduction}
In the big data era, the high volume of datasets will improve the accuracy of learning networks.
However, the ever-increasing dimension of data samples and the size of datasets introduce new challenges to the centralized in-cloud learning networks.
For example, the in-cloud learning networks will experience accuracy degradation since the limited bandwidth in current networking infrastructure may not satisfy the demands for transmitting the \mbox{high-volume} datasets to the central cloud \cite{Konecny2015}.
The central cloud and terminals have the copies of datasets.
Hence, there is a privacy concern as long as either the terminals or central cloud is hacked by malfunctioning users, e.g., the unexpected leakage of photos of celebrities from iCloud in 2014\footnote{\url{https://en.wikipedia.org/wiki/ICloud_leaks_of_celebrity_photos}}.

The soaring demands on bandwidth and privacy concerns cause a shift of research focus from the centralized in-cloud learning networks to the distributed on-device learning networks.
Moreover, the on-device learning networks also benefit from the improvement of storage volume and computational power of mobile terminals.
The objective of on-device learning networks is to obtain the optimal model parameter, which provides an optimal mapping between the data samples and labels of a dataset.
For example, the optimal model parameter is used to map bits of an image to the number in the image in the MNIST dataset\footnote{\url{http://yann.lecun.com/exdb/mnist/}}.

\begin{figure}[htb]
\centering
\subfigure[A typical FLN with $K$ terminals.]{\includegraphics[height = 2 in]{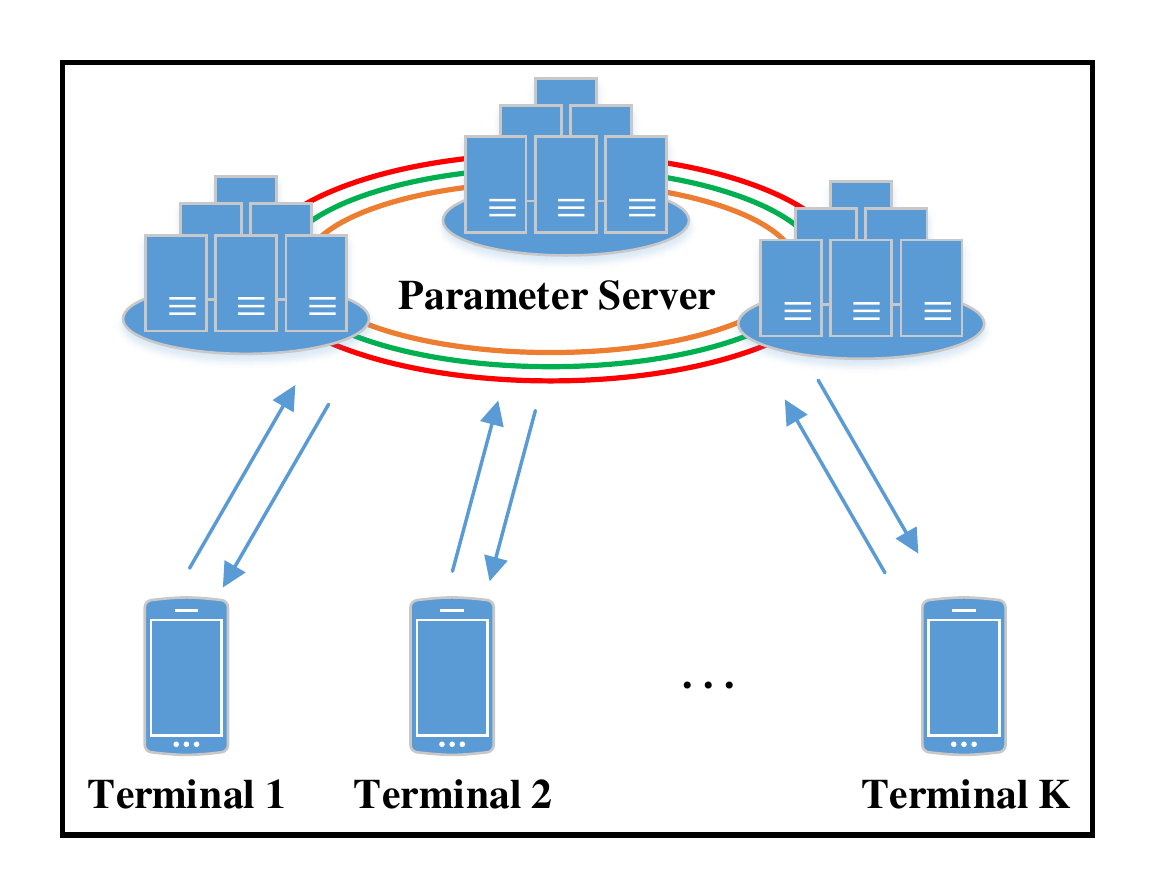}\label{fg:001:a}}
\hspace{0.2 cm}
\subfigure[A typical DLN with five terminals.]{\includegraphics[height = 2 in]{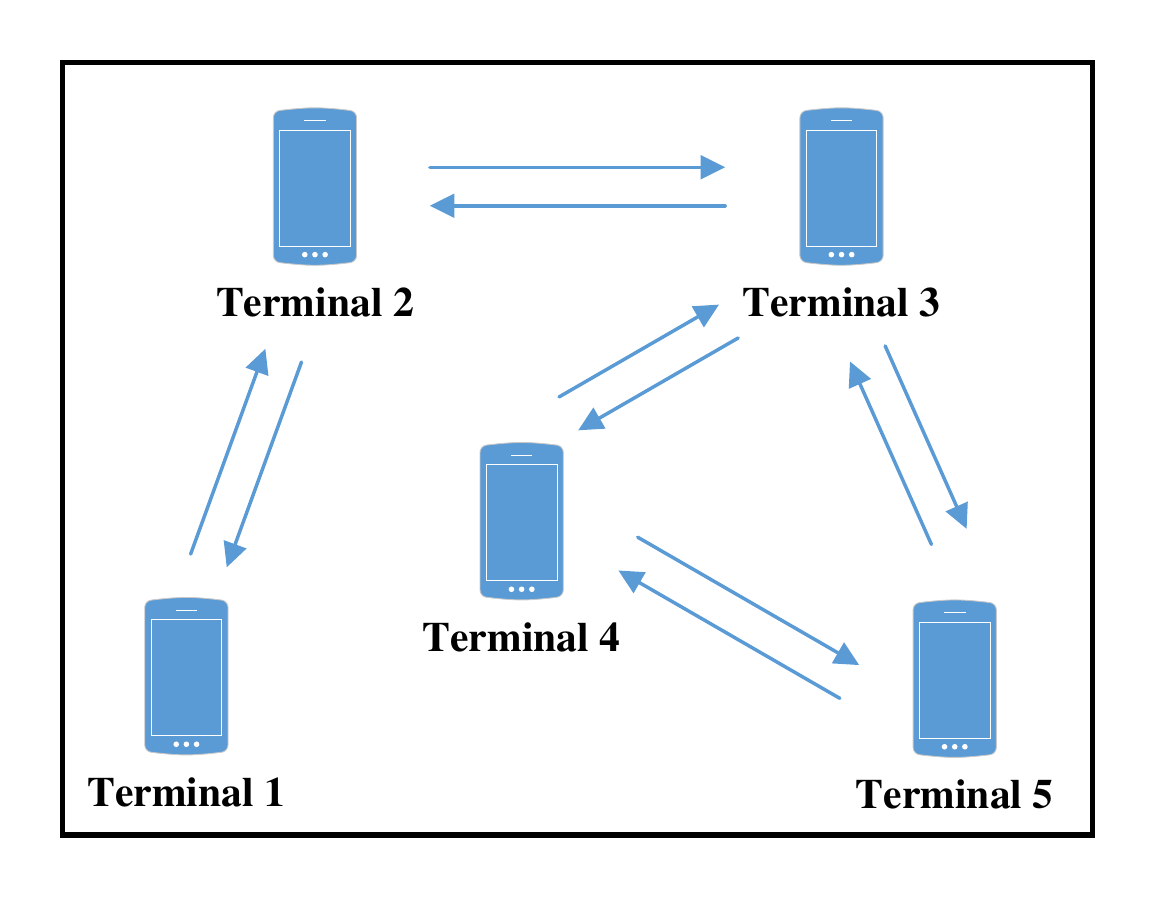}\label{fg:001:b}}
\subfigure[Comparisons of in-cloud learning networks with FLNs and DLNs.]{\includegraphics[height = 1.6 in]{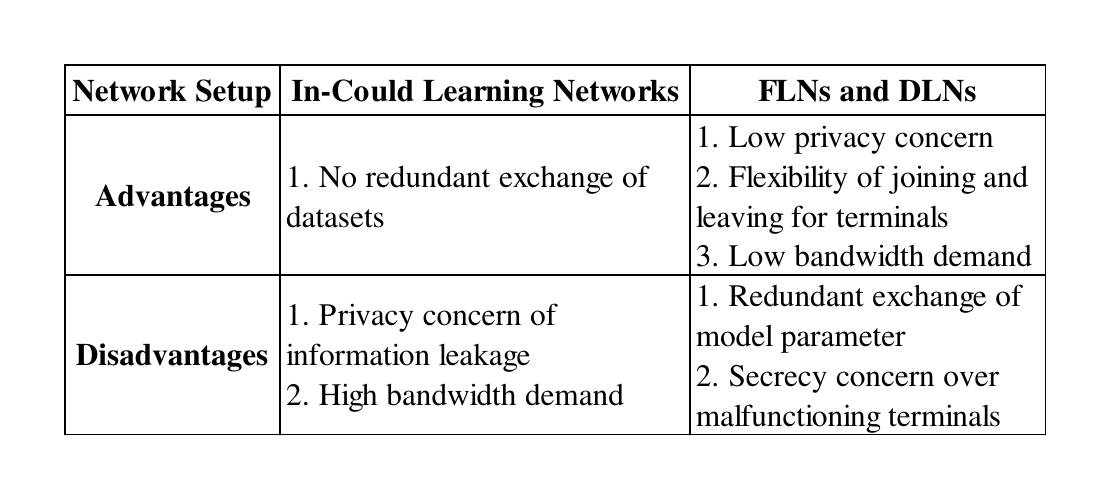}\label{fg:001:c}}
\caption{An illustration of FLNs and DLNs, and the comparisons with in-cloud learning networks.}
\end{figure}

Among the  on-device learning networks, the \mbox{Federated-Learning} networks (FLNs) \cite{Konecny2015} and decentralized-learning networks (DLNs) \cite{Ying2018} are the two categories of promising candidates as shown in Figs. \ref{fg:001:a} and \ref{fg:001:b}, respectively.
As an on-device learning network recently proposed by Google, the FLN consists of a parameter server and multiple terminals as shown in Fig. \ref{fg:001:a}.
In the FLNs, the terminals perform the distributed parallel computation based on the local datasets.
The parameter server updates the model parameter based on the results of the terminals.
Due to the parallel computing, the FLNs converge as fast as the centralized counterpart: the \mbox{in-cloud} learning networks.
When the parameter server experiences unprecedented outage\footnote{The outage events include the energy outage at the location of server, the hardware failure of server, etc.}, the DLN becomes a promising on-device learning network since the DLN does not require a central parameter server \cite{Ying2018}.
In the DLNs, the terminals need to communicate with the neighbors to exchange the distributed training results.
Hereinafter, two terminals are neighbors when they have one-hop connection with each other.
For example, the first and the second terminals are neighbors to each other in Fig. \ref{fg:001:b}.
When each terminal is the neighbor of the remaining terminals, the DLN is called \mbox{fully-connected}; otherwise, it is called partially-connected.
{\color{black}
The advantages and disadvantages of in-cloud learning networks and on-device learning networks are summarized in the table of Fig. \ref{fg:001:c}.}

%


The FLNs and DLNs have several desirable characteristics.
For example, the datasets are kept at the terminals such that the FLNs and DLNs do not have multiple copies of the privacy-related datasets.
Hence, the probability of privacy leakage is reduced.
Another desirable characteristic is that the terminals are flexible to join and leave the FLNs (or DLNs).
In order to achieve these desirable characteristics, several challenges need to be solved such as the uneven distribution of training datasets, redundant exchanging of model parameter and secrecy over malfunctioning terminals \cite{Konecny2015, NIPS2017_6617, chen2017}.
Among these challenges, the secrecy over malfunctioning terminals is the most important one since failure in preserving secrecy degrades the accuracy of FLNs and DLNs.
However, the current literature on the FLNs (or DLNs) mostly assume that the terminals can reliably upload (or exchange) the local results to the parameter server (or neighbors).
Therefore, they cannot be used to secure the FLNs and DLNs over malfunctioning terminals (see \cite{Konecny2015, NIPS2017_6617, chen2017} and references therein).

Several recent research works \cite{NIPS2017_6617, chen2017, pmlr-v80-yin18a, pmlr-v80-mhamdi18a, Chen2018, Litobepublished, Alistarh2018,  Su2016, Sundaramtobepublished} have reported that secure learning algorithms can be used to protect the security of FLNs and DLNs with multiple Byzantine adversaries.
{\color{black}In the FLNs and DLNs, the Byzantine adversaries are the malfunctioning terminals, that can 1) obtain the full malicious action space based on the full knowledge of the networks; and 2) choose arbitrarily bad action from the full malicious action space to compromise the prediction accuracy of the networks \cite{Lamport1982}.
Since the action space of any other malfunctioning terminals is a subset of the full malicious action space, the Byzantine adversaries are considered as the \mbox{worst-case} malfunctioning terminals.
When a learning algorithm preserves secrecy over the Byzantine adversaries, it is robust to the full malicious action space of the networks.
Thus, the learning algorithm is robust to any other malfunctioning terminals.}
Hereinafter, we focus on reviewing the state-of-the-art research progresses on secure learning algorithms in the FLNs and DLNs with Byzantine adversaries.
Our contributions are summarized as follows.
\begin{itemize}
  \item In the FLNs with Byzantine adversaries, we classify the current secure Federated-Learning algorithms (SFLAs) into four categories:
      1) aggregation rule based SFLAs;
      2) preprocess based SFLAs;
      3) model based SFLAs;
      and 4) adversarial detection based SFLAs.
      We also provide qualitative comparisons of current SFLAs.
  \item Since the design of secure decentralized-learning algorithms (SDLAs) is at the early stage, few works investigated the SDLAs.
      Therefore, we review several exemplary works on SDLAs.
  \item We discuss several future research directions for secure learning algorithms in the FLNs and DLNs with Byzantine adversaries.
\end{itemize}
To the best of the authors' knowledge, this is the first comprehensive overview on the secure learning algorithms in the on-device learning networks with Byzantine adversaries.

\section{On-Device Learning Networks With Byzantine Adversaries: Basics}
\subsection{Functions of Devices in the On-Device Learning Networks}
In the on-device learning networks with Byzantine adversaries, secure learning algorithms have two objectives: convergence and optimality \cite{NIPS2017_6617, chen2017, pmlr-v80-yin18a, pmlr-v80-mhamdi18a}.
In order to achieve both goals, the following functions need to be implemented  at different devices of the FLNs and DLNs.

\begin{itemize}
\item \textbf{Parameter server:} In the FLNs, the updating and broadcasting of model parameter are performed at the parameter server.
\item \textbf{Reliable terminals:}
    In the FLNs, the terminals upload the local gradients to the parameter server \cite{NIPS2017_6617, chen2017, Litobepublished, pmlr-v80-mhamdi18a, pmlr-v80-yin18a, Chen2018, Alistarh2018}.
    In the DLNs, the terminals exchange the local model parameter with the neighbors \cite{Su2016, Sundaramtobepublished}.
\item \textbf{Byzantine adversaries:} In the FLNs and DLNs, the Byzantine adversaries perform arbitrarily bad actions.
     Different Byzantine adversaries can collude to compromise the prediction accuracy of learning algorithms  \cite{NIPS2017_6617, chen2017,  pmlr-v80-mhamdi18a}.
\end{itemize}


\subsection{Description of Federated-Learning Networks}
\begin{figure}[tb]
\centering
\subfigure[The $\br{Q+1}$-th to the $K$-th terminals are Byzantine adversaries.]{\includegraphics[height = 2 in]{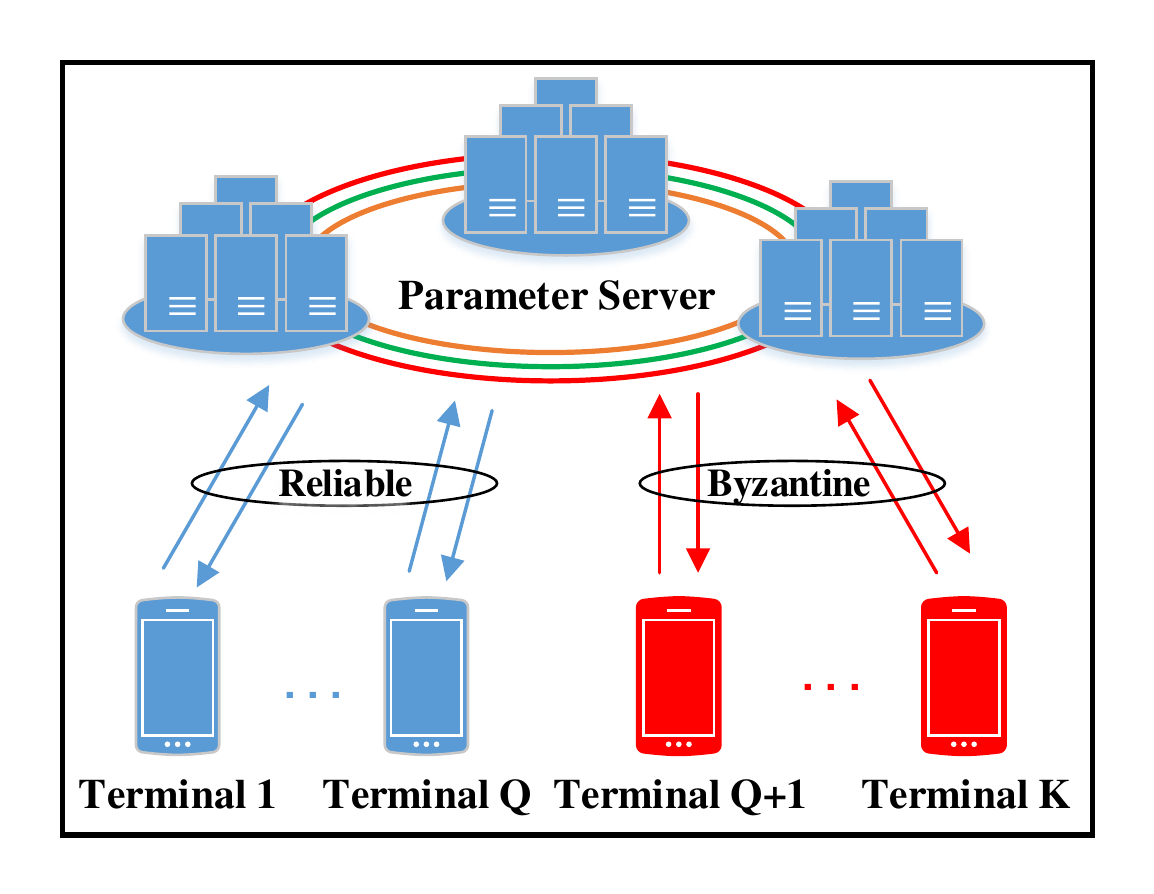}\label{fg:002:a}}
\hspace{0.2 cm}
\subfigure[The implementation procedures of a typical SFLA with Byzantine adversaries.]{\includegraphics[height = 2 in]{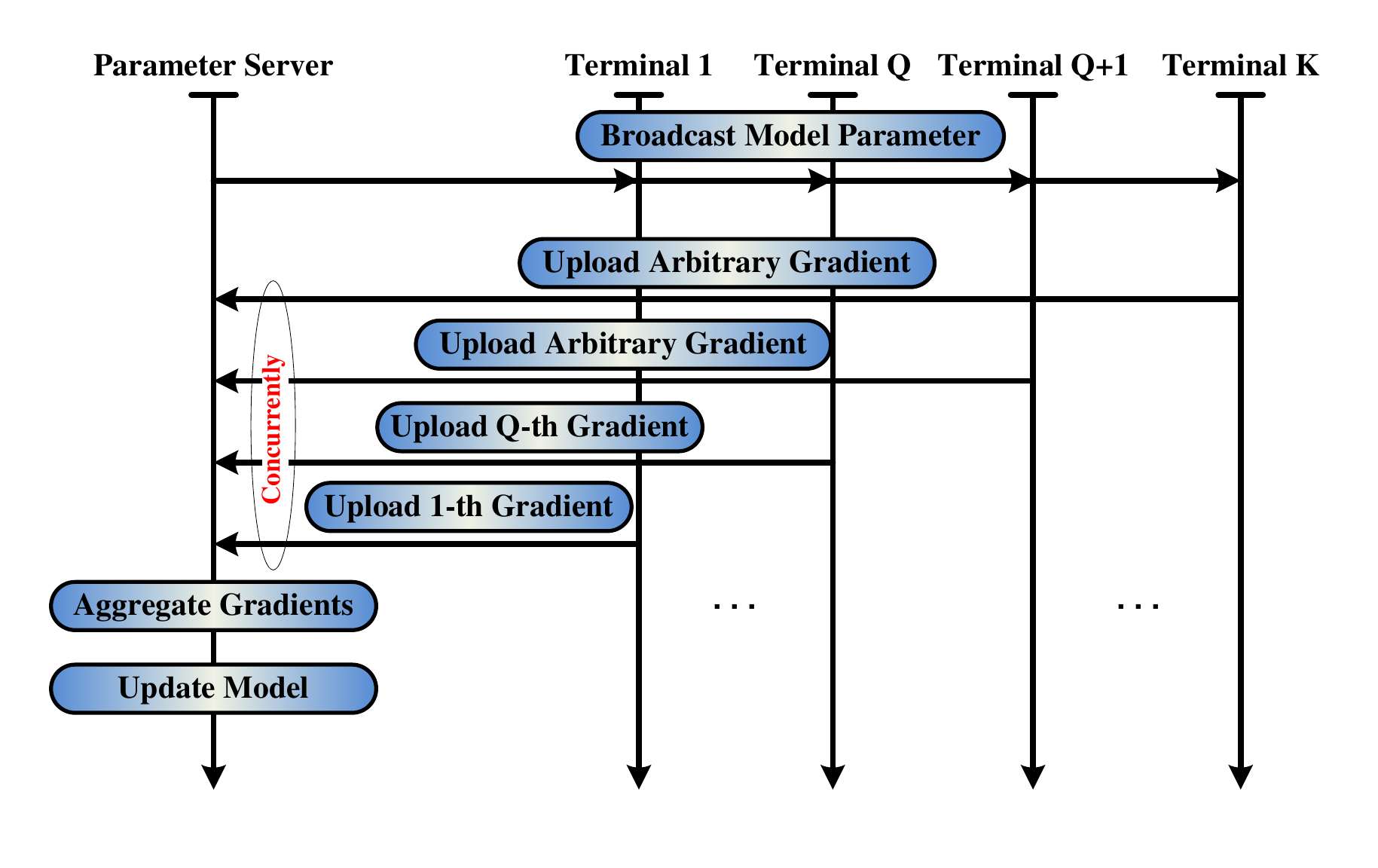}\label{fg:002:b}}
\caption{General setup of FLNs and SFLAs.}
\end{figure}

We consider an FLN with a parameter server and $K$ terminals, where the first $Q$ terminals are reliable and the remaining $K-Q$ terminals are Byzantine adversaries as shown in Fig. \ref{fg:002:a}.
Let $\bm z_k$ denote the randomly-distributed data at the $k$-th terminal.
The distribution of $\bm z_k$ is unknown, and the distributions of data at different terminals can be different.
Let ${\hat f}_k\br{\bm w} = \mathds{E}_{\bm z_k}\sq{ f\br{\bm w; \bm z_k} }$ denote the average loss function with respect to the model parameter $\bm w$ at the $k$-th terminal.
Here, the loss function $f\br{\bm w; \bm z_k}$ quantifies the prediction accuracy associated with the model parameter $\bm w$ at the $k$-th terminal.
Leveraging the parallel computation of terminals, SFLAs in the FLN can obtain the optimal model parameter $\bm w^*$ that minimizes the sum-of-average-loss-function (SoALF) $\sum\nolimits_{k = 1}^K {\hat f}_k\br{\bm w}$ without impairment of $K-Q$ Byzantine adversaries.

The detailed procedures of a typical SFLA in the FLN with Byzantine adversaries are shown in Fig. \ref{fg:002:b}.
At the beginning of $\br{\tau+1}$-th slot, the parameter server broadcasts the global model parameter $\bm w^{\tau}$ to all the $K$ terminals.
After receiving the model parameter $\bm w^{\tau}$, the reliable terminals calculate the local gradient of average loss function $\nabla \hat f_k\br{\bm w^{\tau}}$ based on the data sample (or mini-batch of samples\footnote{Mini-batch of samples is a subset of data samples in a dataset.}) $\bm z_k^{\tau}$.
The Byzantine adversaries generate harmful local gradients.
Then, the terminals synchronously upload the local gradients to the parameter server.
At last, the parameter server obtains a global gradient via the aggregation rule.
Here, an aggregation rule defines a method for the parameter server to obtain the global gradient based on the uploaded local gradients.
Using the global gradient, the parameter server updates the model parameter for the next iteration until the convergence of SoALF.

\subsection{Description of Decentralized-Learning Networks}
\begin{figure}[tb]
\centering
\subfigure[The $2$-nd and $4$-th terminals are  Byzantine adversaries.]{\includegraphics[height = 2 in]{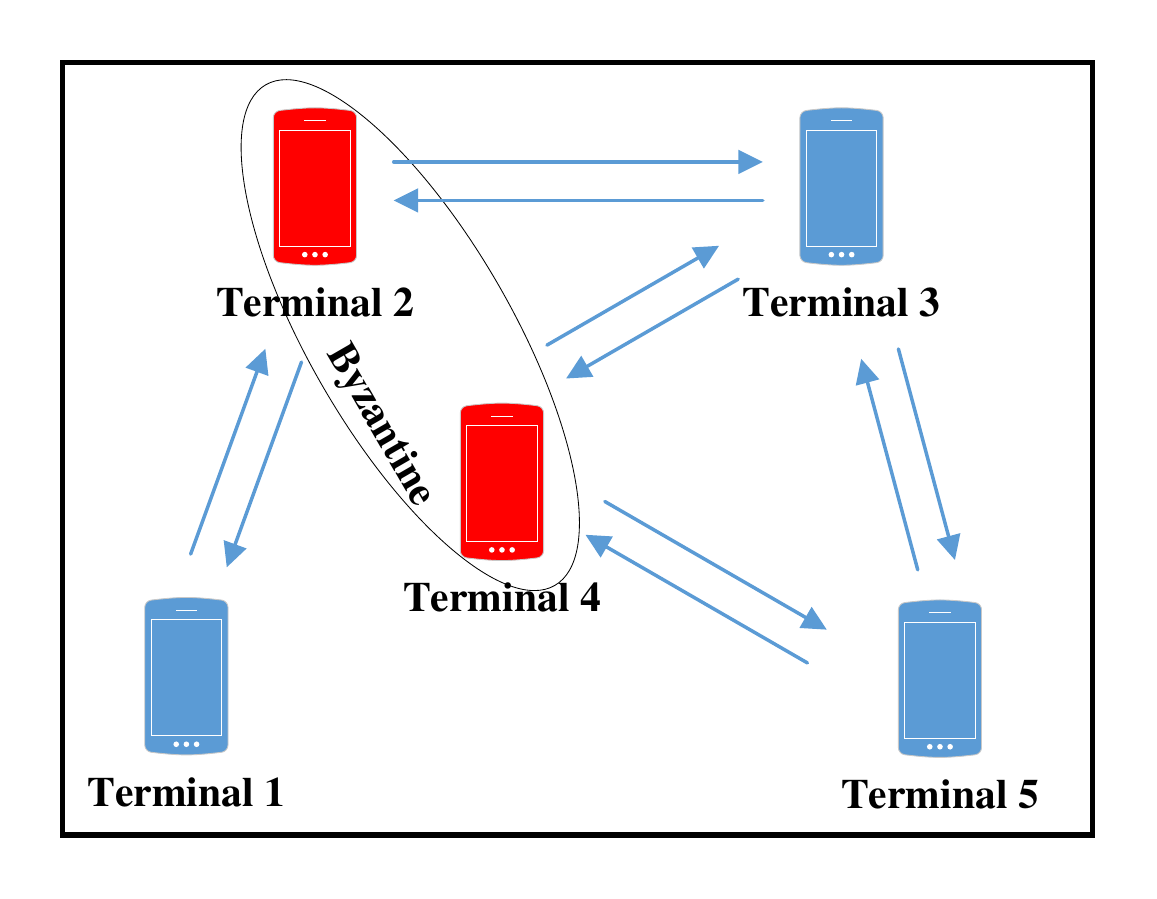}\label{fg:003:a}}
\hspace{0.2 cm}
\subfigure[The implementation procedures of a typical SDLA with Byzantine adversaries.]{\includegraphics[height = 2 in]{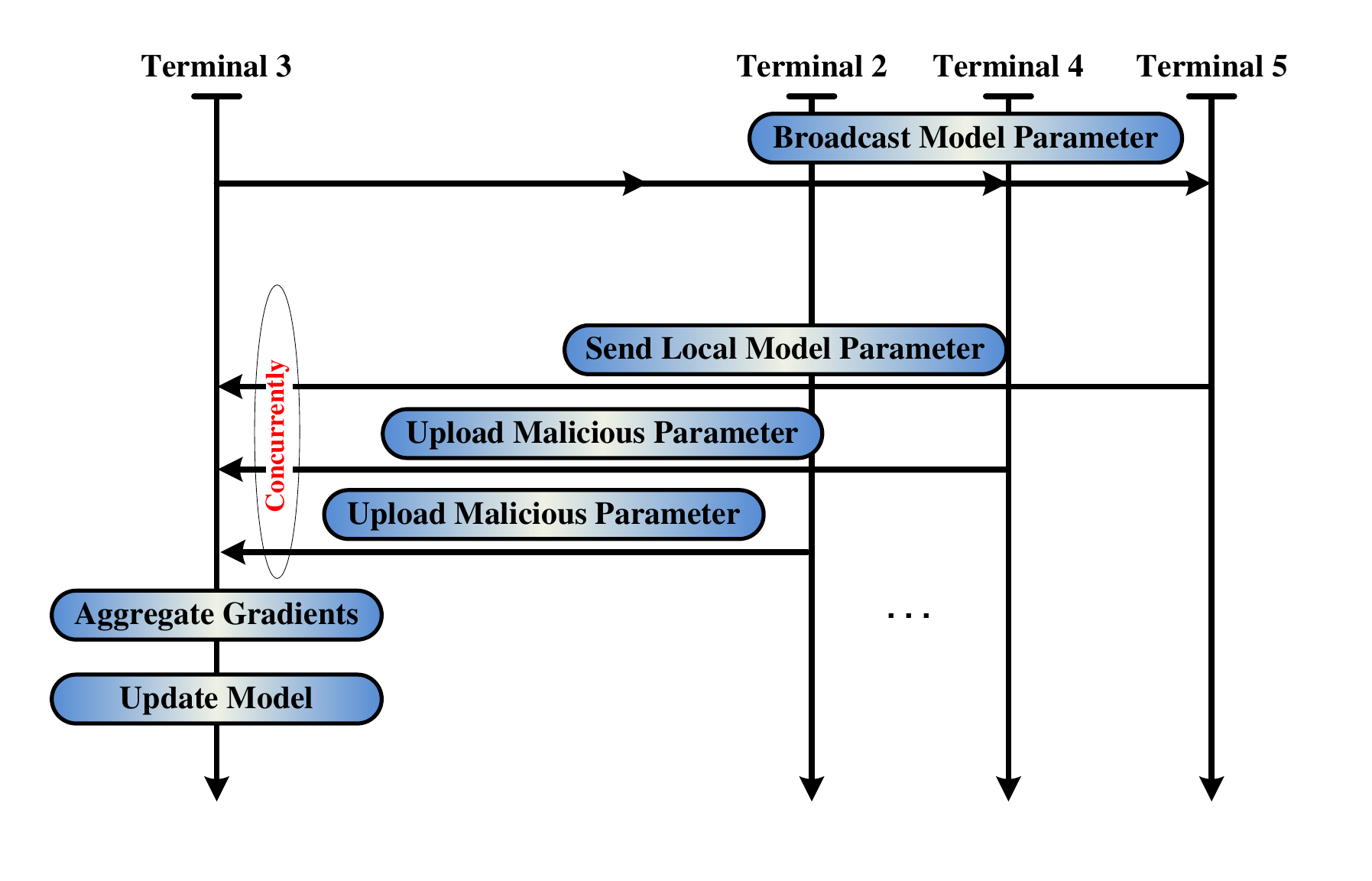}\label{fg:003:b}}
\caption{General setup of DLNs and SDLAs.}
\end{figure}

We consider a DLN with $K = 5$ terminals.
As shown in Fig. \ref{fg:003:a}, the  second and the fourth terminals are Byzantine adversaries among the five terminals.
The first, the third and the fifth terminals are reliable terminals (i.e., $Q = 2$).
Different from the FLNs, the terminals in the DLN obtain the optimal model parameter $\bm w^*$ that minimizes the SoALF as $\sum\nolimits_{k = 1}^5 {\hat f}_k\br{\bm w}$ in a decentralized way.
During the operation of SDLA, each terminal exchanges the local model parameter with the neighbor terminals.
We take the third terminal as an example, and present the detailed procedures of a typical SDLA in the DLN with Byzantine adversaries as shown in Fig. \ref{fg:003:b}.

At the start of the $\br{\tau+1}$-th slot, the third reliable terminal computes the value of the third average loss function $\hat f_3\br{\bm w_3^{\tau}}$ based on the local model parameter $\bm w_3^{\tau}$.
Then, the third reliable terminal updates the local model parameter and obtain $\bm w_3^{\tau+1}$.
Finally, the third reliable terminal broadcasts the local model parameter $\bm w_3^{\tau+1}$ to the neighbor terminals, e.g., the second, the fourth and the fifth terminals.
The Byzantine adversaries broadcast arbitrary adversarial model parameter to their neighbor terminals.
All terminals repeat the above process until the SoALF converges.

{\color{black}
\subsection{Stochastic Gradient, Batch Gradient and Mini-Batch Gradient}
Since the distribution of data $\bm z_k$ is unknown, it is challenging to obtain the exact local gradient $\nabla \hat f_k\br{\bm w}$ due to the high-dimensional integration over the unknown data $\bm z_k$.
Therefore, several estimators of local gradient are obtained based on the ways to use data samples of terminals.
The batch gradient is obtained as $\nabla \hat f_k\br{\bm w^{\tau}} \approx \frac{1}{\ab{\bm z_k^{\tau}}}\sum\nolimits_{i=1}^{\ab{\bm z_k^{\tau}}} \nabla f\br{\bm w^\tau; \bm z_{k_i}^{\tau}}$ where $\ab{\bm z_k^{\tau}}$ and $\bm z_{k_i}^{\tau}$ respectively denote the size of data samples and the $i$-th data sample of the $k$-th terminal at the beginning of $(\tau+1)$-th slot.
Since the batch gradient needs to evaluate the gradient of loss function over all data samples at the $k$-th terminal, the computational cost at the $k$-th terminals increases as the size of data samples.
By sacrificing the convergence rate, the computational cost can be reduced via the stochastic gradient.
The stochastic gradient is obtained by calculating the gradient of loss function over one randomly-drawn data sample as $\nabla \hat f_k\br{\bm w^{\tau}} \approx \nabla f\br{\bm w^\tau; \bm z_{k_i}^{\tau}}$.
In order to balance the computational cost and convergence rate, the mini-batch gradient can be used by evaluating the gradient of loss function over a mini-batch data samples at the $k$-th terminal as $\nabla \hat f_k\br{\bm w^{\tau}} \approx \frac{1}{L}\sum\nolimits_{i=1}^L\nabla f\br{\bm w^\tau; \bm z_{k_i}^{\tau}}$ with $L$ as the size of mini-batch data samples.
}

\section{Prevalent Secure Distributed Learning Algorithms}
\subsection{Aggregation Rule Based Algorithms}

Aggregation of gradients is the important step for an SFLA in the FLNs with Byzantine adversaries.
In their seminal work \cite{NIPS2017_6617}, Blanchard \emph{et al.} reported that the \mbox{mean-value} aggregation rule outputs a sequence of global gradients that can be arbitrarily biased by one Byzantine adversary.
Hence, the learning algorithms with \mbox{mean-value} aggregation rule do not converge or converge to an ineffective model parameter.
{\color{black}
When the fraction of Byzantine adversaries is less than $50\%$, Blanchard \emph{et al.} proposed a secure aggregation rule named as Krum \cite{NIPS2017_6617}.
The objective of Krum is to approximate the optimal global gradient, which is defined as the average of local gradients when no Byzantine adversary exists.
In each iteration, Krum selects one of the local gradients as the global gradient.
Since selected global gradient has the smallest sum Euclidean distance to its first $Q-2$ closest local gradients, the Euclidean distance between the selected global gradient and optimal global gradient is bounded.
Therefore, the impact of Byzantine adversaries is mitigated.}
Moreover, Blanchard \emph{et al.}  also demonstrated that the selected sequence of global gradients almost-surely converges \cite{NIPS2017_6617}.

Compared with the mean value of a sequence, the median gives a reliable estimation when more than half of the sequence is correct.
Hence, several aggregation rules are proposed based on the variants of median \cite{chen2017, pmlr-v80-yin18a}.
For the multi-dimensional local gradients, Chen \emph{et al.}  defined a geometric median (GeoMed) as the minimizer of the sum distances to all local gradients \cite{chen2017}.
With the GeoMed, the parameter server  outputs a global gradient that secures the distributed learning algorithms.
{\color{black}
When the convex SoALF is used, Chen \emph{et al.}  proved that the SFLAs with GeoMed aggregation rule are Byzantine-robust when the number of Byzantine adversaries is less than the number of reliable terminals \cite{chen2017}.}
In addition, Chen \emph{et al.} also quantified the convergence rate of each iteration \cite{chen2017}.

{\color{black}
When the fraction of Byzantine adversaries is less than $50\%$, Yin \emph{et al.}  proposed two aggregation rules \cite{pmlr-v80-yin18a}: component-wise median (CwMed) and component-wise trimmed mean (CwTM).
The CwMed constructs a global gradient, where each entry is the median of entries in the local gradients with the same coordinate \cite[Def. 1]{pmlr-v80-yin18a}.
With CwMed, each entry of global gradient is not polluted by the Byzantine adversaries.
Different from CwMed, the CwTM first \mbox{component-wisely} removes the largest and smallest fraction of entries in the local gradients \cite[Def. 2]{pmlr-v80-yin18a}.
Then, the CwTM constructs a global gradient, where each entry is the mean value of remaining entries of local gradients.
The CwTM needs to know the number of Byzantine adversaries such that the fault entries of Byzantine adversaries are removed with high probability.
Yin \emph{et al.} also quantified the convergence rate of SFLAs with CwMed or CwTM aggregation rules for strongly convex, non-strongly convex and smooth non-convex SoALFs \cite{pmlr-v80-yin18a}.}

Since the GeoMed aggregation rule requires solving an convex optimization to obtain the global gradients, the computational complexity exponentially increases with the dimension of model parameter.
Since Krum, CwMed and CwTM only include linear algebraic operations to obtain the model parameter, these three aggregation rules have lower computational complexity than GeoMed.

\subsection{Preprocessing Based Algorithms}
In addition to the aggregation rule, several preprocessing methods utilizing the local gradients are investigated to design the SFLAs \cite{Chen2018, pmlr-v80-mhamdi18a}.
For example, Chen \emph{et al.} proposed a preprocess based SFLA \cite{pmlr-v80-mhamdi18a}, known as DRACO, in the FLNs.
{\color{black}
DRACO leverages the coding theory to remove the negative effect (i.e., degradation of prediction accuracy) of Byzantine adversaries.
From the information-theoretical perspective, the optimal gradient can be recovered when the reliable terminals report sufficient information to the parameter server.
Specifically, Chen \emph{et al.} demonstrated that DRACO secures the learning algorithms when the number of Byzantine adversaries is less than half of the average number of gradients based on a data sample (i.e., redundancy ratio is higher than $50\%$) \cite{pmlr-v80-mhamdi18a}.
Moreover, the authors illustrated two practical coding schemes to implement DRACO.
Since DRACO secures the learning algorithms via redundant storage of data samples, DRACO does not impose limitation on the fraction of Byzantine adversaries and can secure the traditional learning algorithms.
Besides, the convergence behavior of DRACO highly depends on the used learning procedures.}

\mbox{El Mhamdi} \emph{et al.} proposed a preprocess method \cite{Chen2018}:  Bulyan.
Bulyan uses Krum \cite{NIPS2017_6617} to obtain a subset of uploaded local gradients.
The parameter server constructs a global gradient by taking the component-wise average to the refined subset of local gradients.
We observe that Bulyan introduces extra linear operations to Krum.
Hence, Bulyan converges when the convex and non-convex SoALFs are used.
Bulyan guarantees that the prediction accuracy is not affected by the dimension of data samples.
Since Bulyan is a refinement of Krum by removing several unnecessary local gradients, it has a more strict limitation on the number of Byzantine adversaries.
{\color{black}
More specifically, the number of Byzantine adversaries is less than 25\% of the number of terminals in the FLNs.}


\subsection{Model Based Algorithms}
The SFLAs can be designed during the formulation of problem model in the FLNs.
Li \emph{et al.} proposed a problem model based SFLA \cite{Litobepublished}, which is named as Byzantine-robust stochastic aggregation (BRSA) algorithm.
{\color{black}
In the BRSA algorithm, a regularization term is introduced to the SoALF such that the uploaded local gradients of reliable terminals and Byzantine adversaries are discretized into finite set.
Using the discrete local gradients, the BRSA algorithm converges to a suboptimal model parameter.
Since the problem model used by BRSA is secured, the BRSA algorithm converges without the limitation on the fraction of Byzantine adversaries \cite{Litobepublished}.}
For the convex SoALF, Li \emph{et al.} quantitatively analyzed the relation between the accuracy of BRSA algorithm and the number of Byzantine adversaries \cite{Litobepublished}.
Selecting different regularization terms, the convergence rate can be traded for the reduction of gap between the optimal model parameter and convergent model parameter.
Besides, the BRSA algorithm allows that the data samples at different terminals to have different distributions.

\subsection{Adversarial Detection Based Algorithms}
While the previous works mainly focused on designing Byzantine-resilient learning algorithms, an extra step to detect Byzantine adversaries has been included before the aggregation of local gradients.
{\color{black}
When the SoALF is convex and the number of Byzantine adversaries is less than that of reliable terminals, Alistarh \emph{et al.} proposed the Byzantine stochastic gradient descent (SGD) algorithm to detect the Byzantine adversaries  \cite{Alistarh2018}.
The motivation of this two-threshold approach is based on the facts that the reliable terminals can introduce: 1) limited variation of time-average local gradients, and 2) limited fluctuation for the time-average inner products of local gradient and variation of model parameter.
When either one of the facts is violated, the terminal is detected as a Byzantine adversary.
}
For a Byzantine adversary without violating the two thresholds, Alistarh \emph{et al.} demonstrated that the Byzantine adversary cannot compromise the convergence and accuracy of Byzantine SGD algorithm \cite{Alistarh2018}.
Based on the detection of Byzantine adversaries, the number of local gradients is reduced to one such that the computational complexity at the terminals is reduced.
Through theoretical analysis, the proposed detection method achieves the optimal number of gradient exchange between the parameter server and the terminals \cite[Theorems 3.4 and 3.5]{Alistarh2018}.

\subsection{Qualitative Comparisons}
While GeoMed, Byzantine SGD and BRSA are effective for the convex loss functions, Krum, CwTM, CwMed, Bulyan and DRACO work for both convex and non-convex loss functions.
Since the Bulyan is a refinement of Krum, it has more strict limitations than Krum on the number of Byzantine adversaries.
Table \ref{fg:table} presents a coarse-grained comparison among the state-of-the-art SFLAs.
Note that the complexity in Table \ref{fg:table} indicates the computational complexity to obtain the global gradient at the parameter.
The computational complexity scales with the number of Byzantine adversaries, the dimensions of model parameter, and the number of terminals.

\begin{table*}[htbp]\tiny
  \caption{Coarse-Grained Comparison of SFLAs}\label{fg:table}
  \centering
  \begin{tabular}{|c|c|c|c|c|}
    \hline
    \textbf{Scheme} & \textbf{Loss Function} & \textbf{\% of Byzantine Adversaries} & \textbf{Sample Redundancy} & \textbf{Complexity}  \\\hline
     {Krum} \cite{NIPS2017_6617}
     & Convex, non-convex & $< 50\%$  & $\times$
     &  Medium
     \\\hline
     GeoMed \cite{chen2017}
     & Convex             & $< 50\%$  & $\times$
     &  High
     \\\hline
     CwTM \cite{pmlr-v80-yin18a}
     & Convex, non-convex &   {$< 50\%$}      & $\times$
     &  Low$\sim$Medium     \\\hline
     CwMed \cite{pmlr-v80-yin18a}
     & Convex, non-convex &   {$< 50\%$}     & $\times$     &  Low$\sim$Medium     \\\hline
     BRSA \cite{Litobepublished}
     & Convex             &  N/A      & $\times$     &  Low                 \\\hline
     Bulyan \cite{pmlr-v80-mhamdi18a}
     & Convex, non-convex &  $< 25\%$ & $\times$     &  Medium$\sim$High    \\\hline
     DRACO \cite{Chen2018}
     & Convex, non-convex & N/A       & $\checkmark$ &  Low    \\\hline
     Byzantine SGD \cite{Alistarh2018}
     & Convex             & $< 50\%$  & $\times$
     &  Low
     \\\hline
  \end{tabular}%
\end{table*}%

\subsection{A Case Study}
The loss function takes the form of the regularized softmax regression.
We consider there are 10 terminals, among which there are two Byzantine adversaries. All the results are run over MNIST dataset with $60,000$ training samples and $10,000$ test samples.
We consider two types of attacks: 1) reverse attacks; and 2) Gaussian attacks.
In the reverse attacks, the Byzantine adversaries upload the scaled local gradients scaled by a negative value (e.g., $-100$ in our simulations) to the parameter server.
In the Gaussian attacks, the Byzantine adversaries upload local gradients where each entry follows a Gaussian distribution with mean zero and Gaussian distributed variance.
The variance of the inner Gaussian distribution is set as $100$.
The model parameter is updated by a mini-batch SGD method with mini-batch size of $40$.
Hereinafter, the simulations are run for $5,000$ iterations with the learning rate  $0.1/\sqrt{\mbox{Iteration}}$.

\begin{figure}[htb]
\centering
  \subfigure[Classical learning algorithm without attack.]{\includegraphics[width=3 in]{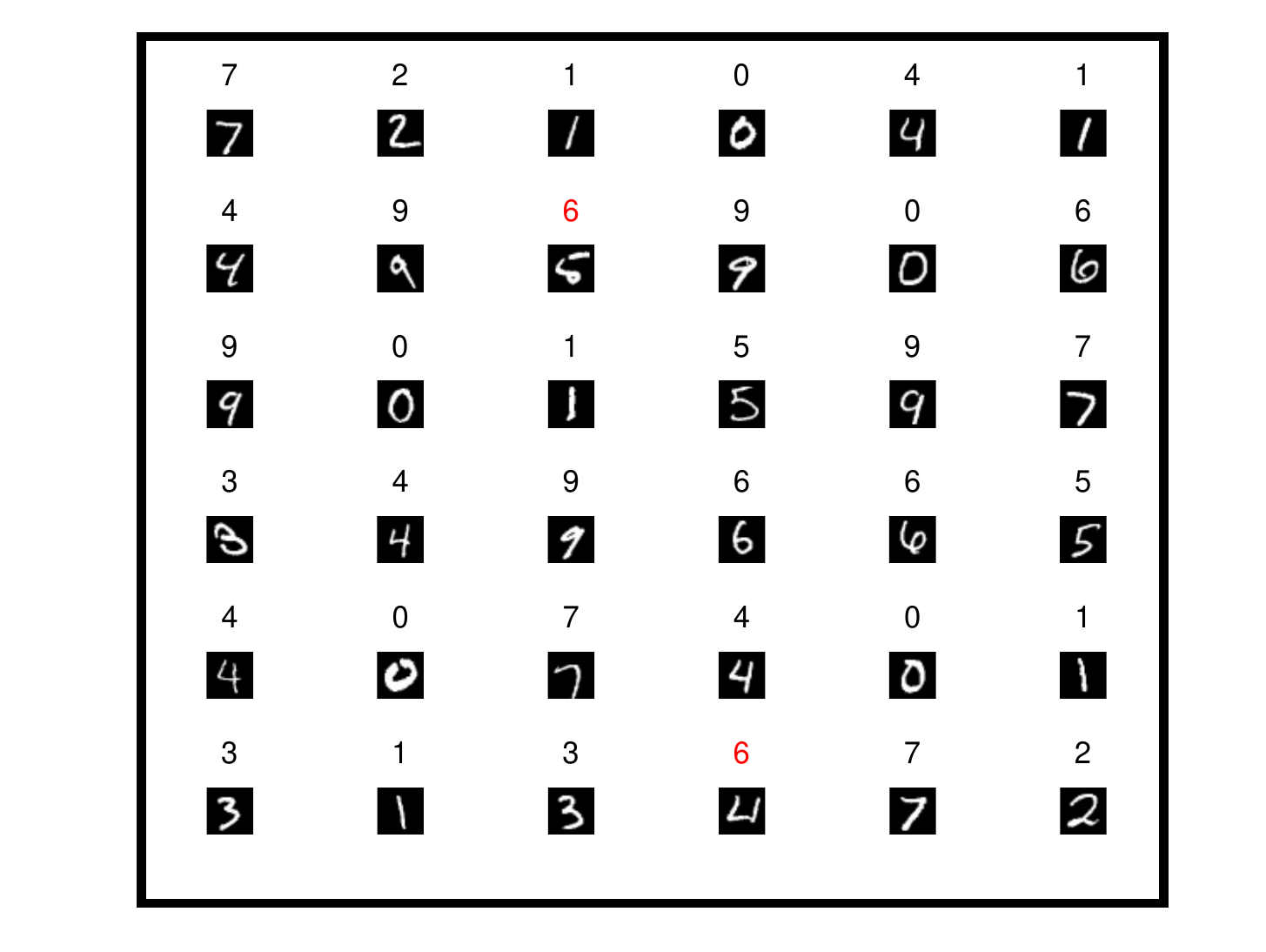}\label{fg:004:a}}
  \subfigure[Classical learning algorithm under reverse attacks.]{\includegraphics[width=3 in]{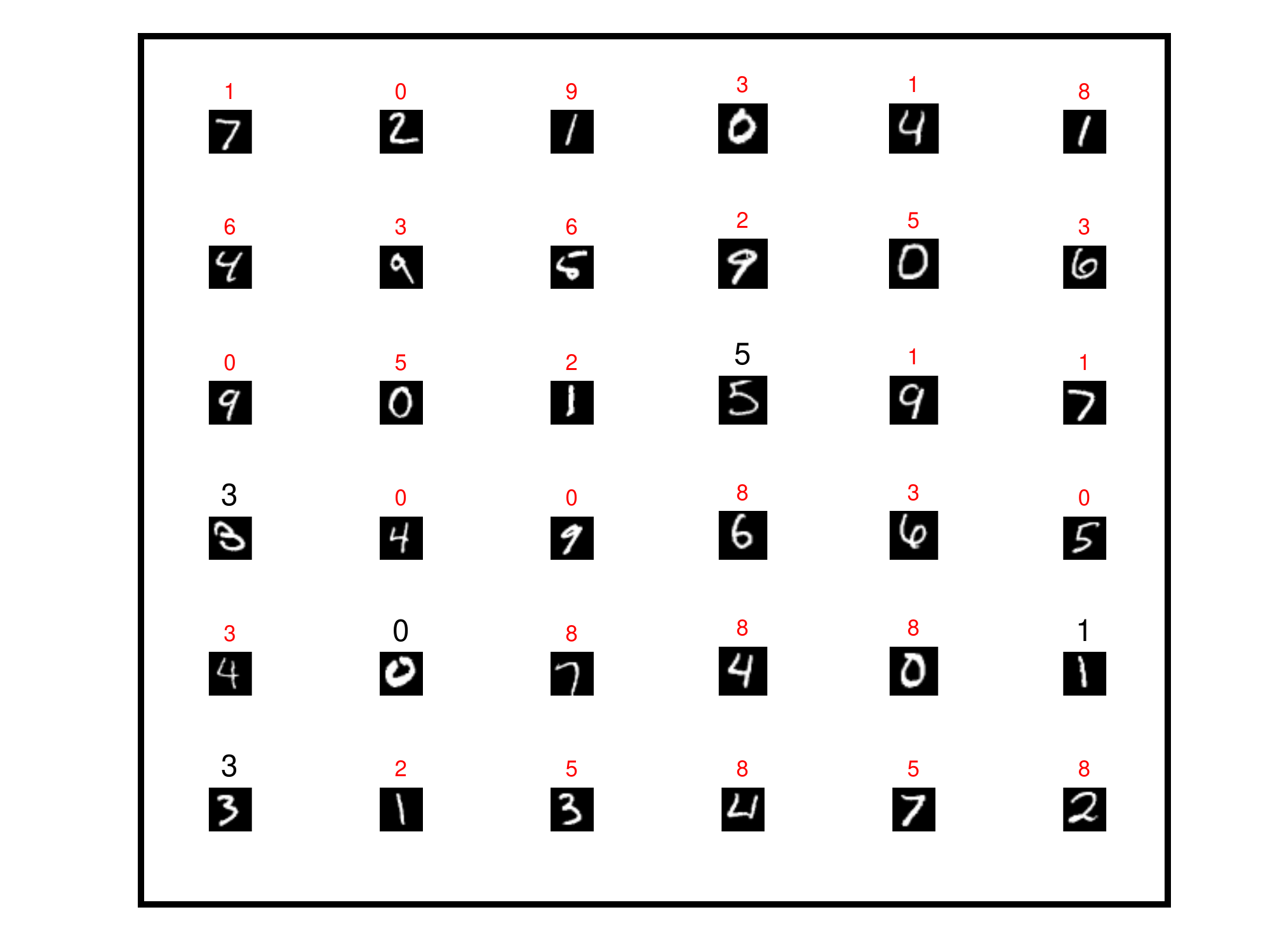}\label{fg:004:b}}
  \subfigure[Classical learning algorithm under Gaussian attacks.]{\includegraphics[width=3 in]{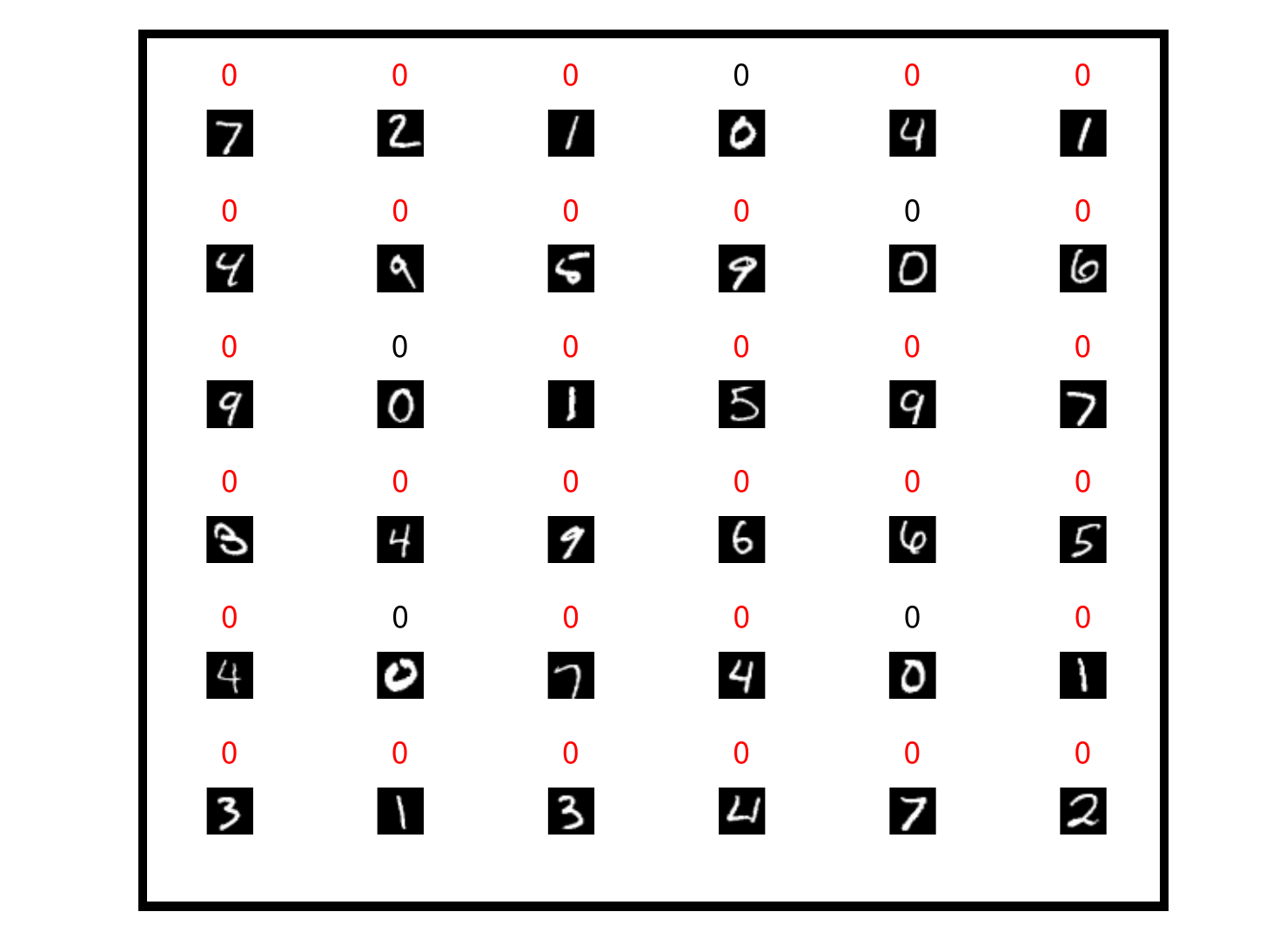}\label{fg:004:c}}
  \caption{An illustration of the impact of Byzantine adversaries on the prediction accuracy. The title of each image shows the predicted value of the image, and the incorrect predicted values are highlighted.}
\end{figure}

Figures \ref{fg:004:a}-\ref{fg:004:c} illustrate that the Byzantine adversaries can compromise the prediction accuracy of the classical learning algorithms.
As shown in Fig. \ref{fg:004:a}, the model parameter obtained by the classical learning algorithms return two errors.
When the Byzantine adversaries are included (e.g., reverse attacks in Fig. \ref{fg:004:b} and Gaussian attacks in Fig. \ref{fg:004:c}), the prediction accuracy of the classical learning algorithms is significantly compromised.

\begin{figure}[htb]
\centering
  \subfigure[Convergence of SFLAs with reverse attacks.]{\includegraphics[width=3 in]{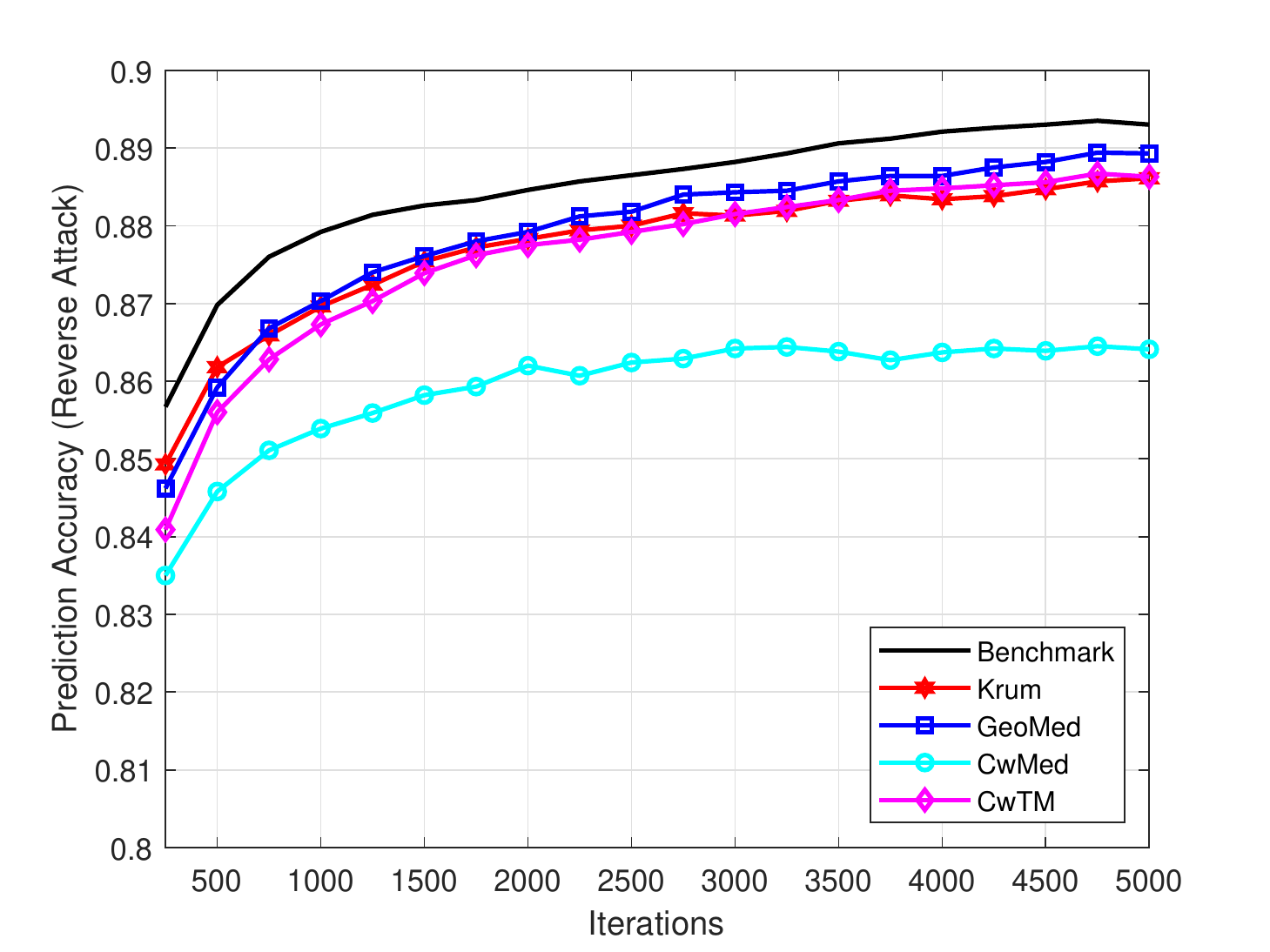}\label{fg:005:a}}
  \subfigure[Convergence of SFLAs with Gaussian attacks.]{\includegraphics[width=3 in]{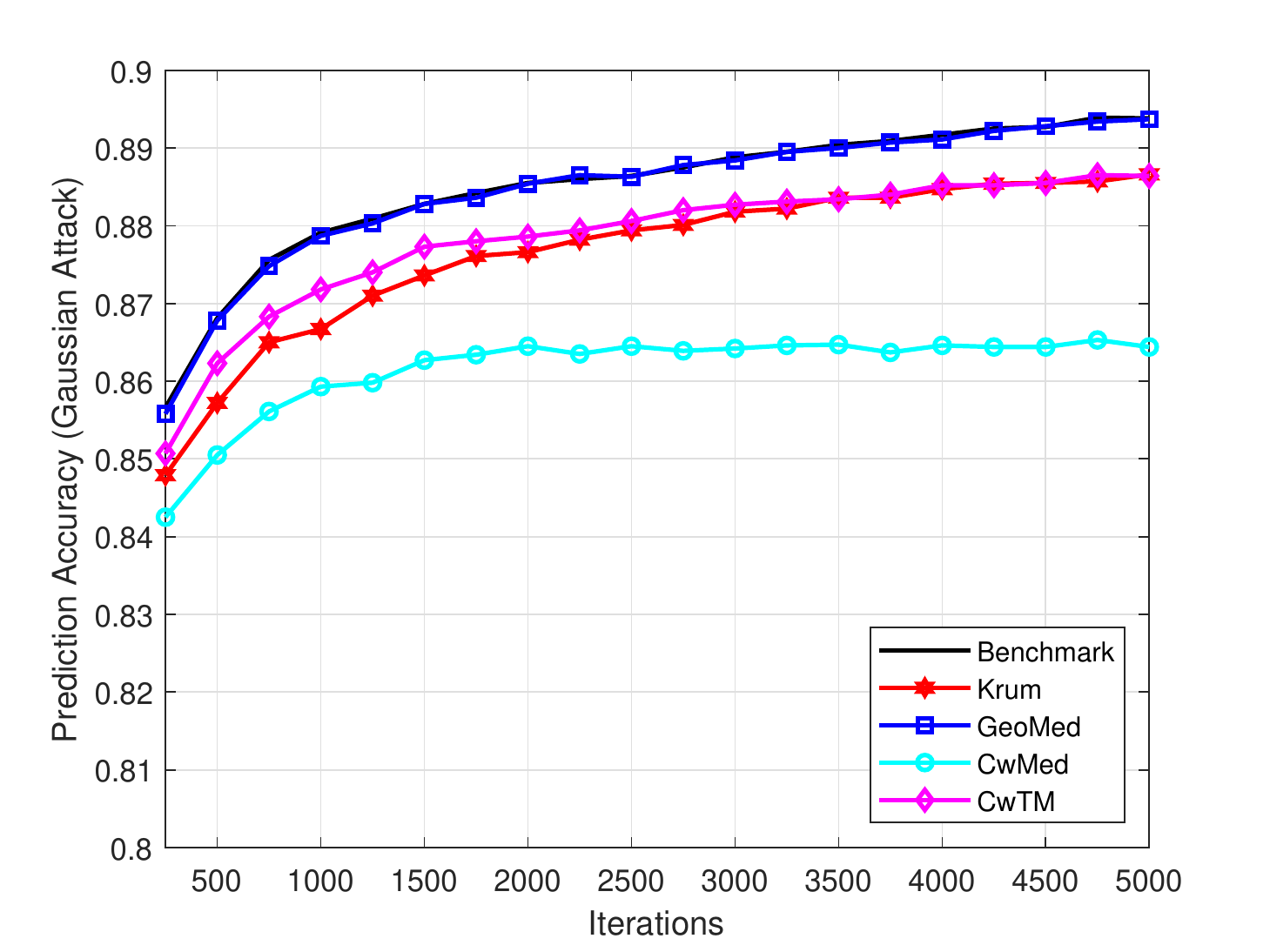}\label{fg:005:b}}
  \caption{Convergence behaviors of SFLAs under different attacks. }
\end{figure}

Figures \ref{fg:005:a} and \ref{fg:005:b} show the accuracy of the four aggregation rules: 1) Krum \cite{NIPS2017_6617}, 2) GeoMed \cite{chen2017}, 3) CwMed \cite{pmlr-v80-yin18a} and 4) CwTM \cite{pmlr-v80-yin18a}.
The benchmark scheme is a classical learning algorithm where model parameter is updated by mini-batch SGD method.
All terminals are reliable in the benchmark scheme.
The Krum algorithm, GeoMed algorithm and CwTM algorithm perform similarly when the reverse attacks are present.
When there are Gaussian attacks, the GeoMed algorithm can obtain a near-optimal model parameter and outperforms the remaining aggregation rules.
However, the GeoMed algorithm requires the parameter server to solve an optimization problem to update the model parameter during each iteration.
The Krum algorithm, CwMed algorithm and CwTM algorithm rely on the simple algebraic manipulations; therefore, the Krum algorithm, CwMed algorithm and CwTM algorithm have lower computational complexities than the GeoMed algorithm.
Hence, we conclude that different aggregation rules provide different tradeoff between the accuracy and the computational complexity.

\section{Prevalent Secure Decentralized Learning Algorithms}
Before proceeding to discuss the SDLAs, we first define the \emph{trim} operation used in DLNs with $K-Q$ Byzantine adversaries.
Performing trim operation to a sequence of scalar values means that the largest $K-Q$ and the smallest $K-Q$ values are removed from the sequence.
In this section, we present two exemplary works for fully-connected DLNs \cite{Su2016} and partially-connected DLNs \cite{Sundaramtobepublished} with a scalar model parameter. In the DLNs, each terminal needs to converge to the same local model parameter, i.e., consensus model parameter.
However, the Byzantine adversaries can broadcast different fake model parameters to their neighbors such that the convergent model parameter is biased from the consensus model parameter.
Moreover, the prediction accuracy of the convergent model parameter is significantly degraded.

In the proposed SDLA for fully-connected DLNs, each terminal exchanges local model parameter and local gradient with the neighbors \cite{Su2016}.
Different from \cite{Su2016}, the SDLA for partially-connected DLNs \cite{Sundaramtobepublished} only requires each terminal to exchange the local gradient with the neighbors.
Therefore, each terminal maintains a local-gradient sequence \cite{Sundaramtobepublished} (or a local-model-parameter sequence and a local-gradient sequence \cite{Su2016}).
In each iteration, each terminal performs the trim operation to the local-gradient sequence \cite{Sundaramtobepublished} (or the local-model-parameter sequence and local-gradient sequence \cite{Su2016}).
Then, each terminal updates the model parameter based on the local-gradient sequence \cite{Sundaramtobepublished} (or the local-model-parameter sequence and local-gradient sequence \cite{Su2016}).

In order to obtain the consensus model parameter, the fully-connected DLNs requires that the reliable terminals are twice larger in number than the Byzantine adversaries.
For the {partially-connected} DLNs, the requirements differ from the connectivity of the terminals.
When there are $K-Q$ Byzantine adversaries in the DLNs, the consensus-based SDLA converges if and only if the partially connected DLNs are $\br{K-Q+1, K-Q+1}$-robust \cite{Sundaramtobepublished}.
When each terminal has at most $K-Q$ Byzantine-adversary neighbors, the consensus-based SDLA converges to the consensual model parameter given the $\br{2K-2Q+1}$-robust DLNs  \cite{Sundaramtobepublished}.

\section{Future Research Directions}

\subsection{Bandwidth Reduction of SFLAs}
In the FLNs, the frequent uploads of gradients from  terminals to  parameter server become inevitable for the SFLAs.
With the high-dimensional gradients and large number of iterations, the bandwidth demand is still high during the learning process of the FLNs.
In order to reduce the bandwidth demand, Chen \emph{et al.} proposed a lazily-aggregated gradient method where the terminals can adaptively skip the gradient calculation and gradient-exchanging  \cite{Chentobepublished}.
Moreover, the authors also theoretically demonstrated that the lazily-aggregated gradient method has the same convergence rate and less communication complexity than the batch gradient descent method.
However, the convergence and optimality of the lazily-aggregated gradient method are unknown when multiple Byzantine adversaries exist in the FLNs.
The \mbox{state-of-the-art} SFLAs require terminals to upload local gradients in each iteration.
The communication complexity becomes a major obstacle to scale up the SFLAs in the FLNs.
In order to reduce the computational complexity of SFLAs, one promising method is to exclude the redundant exchanged gradients of the reliable terminals.
Besides, compressing the exchanged gradients per iteration alleviates the scarcity of bandwidth.
In order to compensate the loss during compression, the SFLAs with gradient compression also need to include  methods such as momentum correction and local gradient clipping.

\subsection{Multi-Task SFLAs}
While the current learning algorithms mainly focus on obtaining a single global model, it is attractive to concurrently compute multiple models when these models are correlated.
The correlation can be induced by similar behaviors of different terminals, such as reposting the similar news in the social media and watching the popular video episodes over internet.
Moreover, the parallel computation of multiple models introduces several benefits such as learning efficiency and prediction accuracy.
Therefore, the multi-task learning algorithms are proposed for the FLNs \cite{Smith2017}.
In the presence of Byzantine adversaries, one potential research direction is to investigate the secure multi-task learning algorithms (SMTLAs) in the FLNs.
The SMTLAs need to converge to a near-optimal model parameter without being falsified by Byzantine adversaries, and also need to preserve the communication expenditure as minimal as possible.
The SMTLAs can be developed based on the aforementioned aggregation rules and regularization term.
However, the convergence property and communication expenditure  remain agnostic when the aggregation rules and regularization term are used.
Moreover, introducing a precoding process for the update of gradients can also secure the multi-task learning algorithms in the FLNs.

\subsection{Stochastic SDLAs}
Since the data samples are locally collected at each terminal in the DLNs, each terminal may have a similar model parameter to its neighbors but retain a small difference \cite{KoppelJune2017}.
As pointed by Koppel \emph{et al.} in \cite{KoppelJune2017}, the proximity constraints among the neighboring terminals are a promising technique to formulate such small differences in the DLNs.
In order to avoid multiple assessments of average  loss functions, the stochastic saddle point algorithm was introduced by allowing each terminal to access its local loss function once in each iteration \cite{KoppelJune2017}.
However, the convergence and optimality of stochastic saddle point algorithm can be compromised by the multiple Byzantine adversaries in the DLNs.
While the proposed algorithms in \cite{Su2016, Sundaramtobepublished} are secure to DLNs with Byzantine adversaries and the scalar model parameter, it is more practical to learn a high-dimensional model parameters due to the ever-increasing volume of datasets.
The regularization term used in \cite{Litobepublished} is promising to handle the Byzantine adversaries in the FLNs, it remains an open problem to design the regularized term in the DLNs with the Byzantine adversaries.
When the heterogeneity of data samples is considered, the design of multi-task SDLAs is also an interesting research problem  in the DLNs.

\section{Conclusions}
Since the current learning algorithms are vulnerable to the Byzantine adversaries, we provided a comprehensive overview of the SFLAs and SDLAs in the FLNs and DLNs, respectively.
The Byzantine adversaries are considered since the Byzantine adversaries can act arbitrarily to compromise the classical learning algorithms.
Therefore, the secure learning algorithms, which are robust over the Byzantine adversaries, can work under any attacks from the terminals.
We presented the signaling-exchange procedures of the secure learning algorithms in both FLNs and DLNs when Byzantine adversaries coexist with the reliable terminals.
Numerous state-of-the-art secure learning algorithms were discussed in terms of the main contributions in the FLNs and DLNs.
Several future research directions were discussed for the secure learning algorithms in the FLNs and DLNs.

\bibliographystyle{IEEEtran}
\bibliography{mach_learn_dyj}

\begin{IEEEbiographynophoto}{Yanjie~Dong~(S'13)}
is currently pursuing a Ph.D. degree with the Department of Electrical and Computer Engineering, The University of British Columbia, Vancouver, Canada.
He was awarded a Four-Year Doctoral Fellowship of The University of British Columbia in 2016.
He received the Graduate Support Initiative Award in 2016, 2017 and 2018.
He was an exemplary reviewer for IEEE Communications Letters and IEEE Transactions on Communications in 2019.
His research interests focus on the protocol design of energy-efficient communications, machine learning and large-scale optimization, and UAV communications.
He served as the Web Committee Chair of GameNets'16.
He also serves/served as a Technical Program Committee member for IEEE ICC'18 and ICC'19, IEEE ICC'18 Workshop on Integrating UAV into 5G, IEEE VTC-Fall'18 and VTC-Spring'19, IEEE WCNC'19, IEEE/CIC ICCC'17 and ICCC'18, IWCMC'18 and 5GWN'19.
\end{IEEEbiographynophoto}

\begin{IEEEbiographynophoto}{Julian~Cheng~(S'96--M'04--SM'13)}
received the B.Eng. degree (Hons.) in electrical engineering from the University of Victoria, Victoria, BC, Canada, in 1995, the M.Sc.(Eng.) degree in mathematics and engineering from Queens University, Kingston, ON, Canada, in 1997, and the Ph.D. degree in electrical engineering from the University of Alberta, Edmonton, AB, Canada, in 2003. He is currently a Full Professor in the School of Engineering, Faculty of Applied Science, The University of British Columbia, Kelowna, BC, Canada. He was with Bell Northern Research and NORTEL Networks. His current research interests include digital communications over fading channels, statistical signal processing for wireless applications, optical wireless communications, and 5G wireless networks. He was the Co-Chair of the 12th Canadian Workshop on Information Theory in 2011, the 28th Biennial Symposium on Communications in 2016, and the 6th EAI International Conference on Game Theory for Networks (GameNets 2016). He currently serves as an Area Editor for the IEEE TRANSACTIONS ON COMMUNICATIONS, and he was a past Associate Editor of the IEEE TRANSACTIONS ON COMMUNICATIONS, the IEEE TRANSACTIONS ON WIRELESS COMMUNICATIONS, the IEEE COMMUNICATIONS LETTERS, and the IEEE ACCESS. Dr. Cheng served as a Guest Editor for a Special Issue of the IEEE JOURNAL ON SELECTED AREAS IN COMMUNICATIONS on Optical Wireless Communications. He is also a Registered Professional Engineer with the Province of British Columbia, Canada. Currently he serves as the President of the Canadian Society of Information Theory.
\end{IEEEbiographynophoto}

\begin{IEEEbiographynophoto}{Md.~Jahangir~Hossain~(S'04--M'08--SM'17)}
received the B.Sc. degree in electrical and electronics engineering from Bangladesh University of Engineering and Technology (BUET), Dhaka, Bangladesh; the M.A.Sc. degree from the University of Victoria, Victoria, BC, Canada, and the Ph.D. degree from the University of British Columbia (UBC), Vancouver, BC, Canada.
He served as a Lecturer at BUET.
He was a Research Fellow with McGill University, Montreal, QC, Canada; the National Institute of Scientific Research, Quebec, QC, Canada; and the Institute for Telecommunications Research, University of South Australia, Mawson Lakes, Australia.
His industrial experiences include a Senior Systems Engineer position with Redline Communications, Markham, ON, Canada, and a Research Intern position with Communication Technology Lab, Intel, Inc., Hillsboro, OR, USA.
He is currently working as an Associate Professor in the School of Engineering, UBC Okanagan campus, Kelowna, BC, Canada.
His research interests include designing spectrally and power-efficient modulation schemes, quality of service issues and resource allocation in wireless networks, and optical wireless communications.
He is serving as an Associate Editor for the IEEE Communications Surveys and Tutorials.
He was also an Editor for the IEEE Transactions on Wireless Communications. Dr. Hossain regularly serves as a member of the Technical Program Committee of the IEEE International Conference on Communications (ICC) and IEEE Global Telecommunications Conference (Globecom).
\end{IEEEbiographynophoto}

\begin{IEEEbiographynophoto}{Victor~C.~M.~Leung~(S'75--M'89--SM'97--F'03)}
is a Distinguished Professor of Computer Science and Software Engineering at Shenzhen University.
He is also an Emeritus Professor of Electrical and Computer Engineering at the University of British Columbia (UBC), where he held the positions of Professor and the TELUS Mobility Research Chair until the end of 2018.
His research is in the broad areas of wireless networks and mobile systems. He has co-authored more than 1200 journal/conference papers and book chapters.
Dr. Leung is serving on the editorial boards of the IEEE Transactions on Green Communications and Networking, IEEE Transactions on Cloud Computing, IEEE Access, IEEE Network, and several other journals.
He received the IEEE Vancouver Section Centennial Award, 2011 UBC Killam Research Prize, 2017 Canadian Award for Telecommunications Research, and 2018 IEEE TGCC Distinguished Technical Achievement Recognition Award. He co-authored papers that won the 2017 IEEE ComSoc Fred W. Ellersick Prize, 2017 IEEE Systems Journal Best Paper Award, and 2018 IEEE CSIM Best Journal Paper Award.
He is a Fellow of IEEE, the Royal Society of Canada, Canadian Academy of Engineering, and Engineering Institute of Canada.
\end{IEEEbiographynophoto}

\end{document}